\documentclass{article}

     \PassOptionsToPackage{numbers, compress}{natbib}

\usepackage[preprint]{neurips_2021}

\usepackage[utf8]{inputenc} 
\usepackage[T1]{fontenc}    
\usepackage{hyperref}       
\usepackage{url}            
\usepackage{booktabs}       
\usepackage{amsfonts}       
\usepackage{nicefrac}       
\usepackage{microtype}      
\usepackage{xcolor}         

\usepackage{times}
\usepackage{epsfig}
\usepackage{graphicx}
\usepackage{amsmath,amssymb,amsthm}
\usepackage{multirow}
\usepackage{wrapfig}
\usepackage{flushend}

\usepackage[ruled,linesnumbered]{algorithm2e}

\title{Symmetric Transformer-based Network for Unsupervised Image Registration}

\author{%
Mingrui Ma, Lei Song, Yuanbo Xu, Guixia Liu \\
Key Laboratory of Symbolic Computation and Knowledge
Engineering of Ministry of Education,\\ College of Computer Science and Technology, Jilin University\\
  \texttt{mamr19,songlei@mails.jlu.edu.cn} \\
  \texttt{liugx,yuanbox@jlu.edu.cn} \\
}

\begin{document}

\maketitle

\begin{abstract}
Medical image registration is a fundamental and critical task in medical image analysis. With the rapid development of deep learning, convolutional neural networks (CNN) have dominated the medical image registration field. Due to the disadvantage of the local receptive field of CNN, some recent registration methods have focused on using transformers for non-local registration. However, the standard Transformer has a vast number of parameters and high computational complexity, which causes Transformer can only be applied at the bottom of the registration models. As a result, only coarse information is available at the lowest resolution, limiting the contribution of Transformer in their models. To address these challenges, we propose a convolution-based efficient multi-head self-attention (CEMSA) block, which reduces the parameters of the traditional Transformer and captures local spatial context information for reducing semantic ambiguity in the attention mechanism. Based on the proposed CEMSA, we present a novel Symmetric Transformer-based model (SymTrans). SymTrans employs the Transformer blocks in the encoder and the decoder respectively to model the long-range spatial cross-image relevance. We apply SymTrans to the displacement field and diffeomorphic registration. Experimental results show that our proposed method achieves state-of-the-art performance in image registration. Our code is publicly available at \url{https://github.com/MingR-Ma/SymTrans}.
\end{abstract}

\section{Introduction}
Medical image registration is the fundamental and crucial branch of many medical image analysis tasks. Deformable medical image registration, a part of the medical image registration, aims to establish the dense and nonlinear correspondence between a pair of images. Traditional image methods formulate image registration as an optimization problem to search for a smooth transformation between the points in the pair of images \cite{2014-spatiallyGuassianSomoothRegsitration,2014-novel-pointMatchingRegsitration}. However, the traditional methods are very time-consuming and require a lot of computing resources because iterative optimization is required every time for a new image pair.

Since recently, with the rapid development of deep learning, convolutional neural networks (CNN) have been applied in many vision tasks and demonstrated the outperformance in many vision tasks \cite{IJCAI-detection,2021-super-resolution-tianchunwei,IJCAI-segmentation}. Compared to the traditional methods in medical image registration, CNN-based methods can improve the registration performance and compute the dense transformation faster once the CNN model train is finished. However, the inherent limitation of the CNN architectures, that is, the local convolution operation (i.e., the local receptive field of CNN), makes the CNN-based methods unable to obtain the long-range spatial relations \cite{2020-CNNLimitation-VIT}. Although some approaches have been proposed to enlarge the local receptive field of CNN, they are still restricted by the kernel size of the convolution \cite{2019-unet++,2015-Unet}. 

The Transformer module that performs well in natural language processing tasks does not have the limitation of local receptive fields.  Benefiting from the non-local receptive field capability of the Transformer, VIT \cite{2020-CNNLimitation-VIT} is the first to apply the Transformer in computer vision (CV), which regards an image as a sequence of patches (i.e., making one image into tokens), achieves the state-of-the-art image recognition results. Recently, many Transformer-based or variant Transformer-based methods have been proposed to model the CV tasks, such as Swin Transformer \cite{SwinTrans} and transU-Net \cite{2021-transunet}.

In medical image registration, the size of the local receptive field of CNN itself will limit the performance of the CNN-based model to establish the correspondence between the same anatomical structures of two images, especially when the same anatomical structure is distant. Based on the Transformer studies in CV, some image registration approaches have utilized the Transformer in their methods. Vit-V-Net \cite{2021-vitV}, as we know, is the first to apply the Transformer in image registration and achieves promising performance. There are also other Transformer-based image registration methods, such as DTN \cite{2021-zhang-DTN} and TransMorph \cite{2021-transmorph}. However, the limited memory and a large number of parameters force them to apply the Transformer at the bottom of their networks, where the coarse feature maps are available. The lowest level resolution information limits the contribution of the Transformer.

To address these issues, we propose an encoder-decoder scheme model consisting of convolutional and Transformer blocks. We present the convolution-based efficient multi-head self-attention (CEMSA), which focuses on capturing local and long-range contextual information. Specifically, we utilize the depth-wise separable convolutional operations to capture the local contextual feature maps and compress the memory and parameters. We use our proposed patch expanding to restore the feature maps from the last CEMSA-based Transfomer encoder to build the symmetric encoder-decoder architecture. Then, the skip connections and the proposed merging operations are used to restore and fuse feature maps in the decoder. Based on these proposed modules, we build the CEMSA-Transformer-based symmetric network (SymTrans). We also introduce a variant model diff-SymTrans to obtain the diffeomorphic deformation field. Qualitative and quantitative evaluation of the experimental results demonstrates the outperformance of the proposed method in image registration.

In summary, the main contributions of this work are following:
\begin{itemize}
\setlength{\itemsep}{0pt}
\setlength{\parsep}{0pt}
\setlength{\parskip}{0pt}
    \item \emph{CEMSA:} We propose an efficient multi-head self-attention mechanism to save memory, reduce parameters, and capture the local relevance.
    \item \emph{An CEMSA-Transformer-based symmetric architecture:} We present a novelty CEMSA-Transformer-based symmetric network, SymTrans, for deformable image registration.
    \item \emph{Displacement and diffeomorphic registration:} We present the two registration fashions, SymTrans, and diff-SymTrans. SymTrans yield the displacement field for registration, and diff-Trans yield the deformation field, ensuring the diffeomorphic properties.
    \item \emph{State-of-the-art results:} We compare SymTrans and diff-SymTrans with three unsupervised learning-based and one wildly used traditional registration approaches. The experimental results demonstrate the state-of-the-art performance.
\end{itemize}


\section{Background}
\subsection{Image Registration}
Deformable image registration aims at establishing spatial correspondence between two images. The registration of a pair of images can be optimized by an energy function. The typical optimization problem is written as:
\begin{equation}
    \begin{aligned}
        \hat{\phi}=\mathop{\arg\min}\limits_{\theta}\mathbb{E}(I_m,I_f,\phi).
    \end{aligned}
    \label{eq:goal}
\end{equation}
In this energy function, $I_m$ and $I_f$ denote the moving and fixed image, respectively. And $\phi$ denotes the deformation field, which indicates the directions and magnitudes of a spatial pixel point's transformation.  $\mathbb{E}$ can be fomulated as:
\begin{equation}
    \begin{aligned}
        \mathbb{E}(I_m,I_f,\phi)=\mathbb{E}_{sim}({I_m}{{\circ}{\phi}},I_f)+{\lambda}\mathbb{R}(\phi),
    \end{aligned}
    \label{eq:energy}
\end{equation}
where $\mathbb{E}_{sim}(\cdot)$ is the similarity metric, $\circ$ is the interpolation operation, and ${I_m}{{\circ}{\phi}}$ is the warped image warped by the deformation field $\phi$. The similarity function is the metric to evaluate the level of alignment between the warped moving image (i.e., ${I_m}{{\circ}{\phi}}$) and the fixed image ${I_f}$. $\mathbb{R}(\cdot)$ is a regularizer that enforces the deformation smooth. $\lambda$ is the hyperparameter to balance the contributions of the similarity and the regularization.

\subsection{Vision Transformer}
A standard Transformer block consists of two components: multi-head self-attention (MSA) and position-wise feed forward module (FFN) \cite{2017-attention}. Let I is an image volume defined in the 3D spatial domain ${\Omega}\subset{\mathcal{R}}^{D{\times}H{\times}W}$. To use the Transformer model the input volume, an image is first divided into N patches, then flattened to sequences of vectors ${I_p}\subset{\mathcal{R}^{N{\times}P^3}}$. The number of patches can be calculated by the formula $N=\frac{D{\times}H{\times}W}{P^3}$, where (\emph{D}, \emph{H}, \emph{W}) is the size of the image, \emph{P} is the size of each patch. Usually, the convolutional operation is utilized to split an image into patch embeddings without overlap \cite{2021-ResT,2020-CNNLimitation-VIT}. After getting the patch embeddings of an image, these embeddings are passed in the MSA. MSA applies the linear operation project the embeddings to the queries, keys, and values (denoted as $\mathbf{{Q}}$, $\mathbf{{K}}$, $\mathbf{{V}}$). Each linear projection set consists of \emph{k} heads, which map the $d_m$ dimensional input into $d_k$ dimensional space. The input sequences to the global relations can be formulated as:
\begin{equation}
    \begin{aligned}
        \text{MSA}(\mathbf{Q},\mathbf{K},\mathbf{V})=\text{Softmax}(\frac{\mathbf{{QK}^T}}{\sqrt{d_k}}){\mathbf{V}}.
    \end{aligned}
    \label{eq:MSA}
\end{equation}
The FFN is utilized to project the output sequence from MSA into a large scale (usually by the factor of 4) dimensional space and then project it to the sequence's original dimensional space. Thus, a Transformer block is completed.

\section{Related Work}
Traditional deformable image registration methods optimize the energy function formulated as Eq. \ref{eq:energy} iteratively for each pair of images. These methods include, Demons \cite{demons}, elastic model \cite{elastic}, and two commonly used methods SyN \cite{SyN} and LDDMM \cite{LDDMM}. These methods, as traditional methods, still face the problem of time-consuming calculations.

Unlike the traditional approaches, CNN-based methods learn the parameters of their model on the training dataset to predict the deformation field between a pair of unseen images. Therefore, CNN-based methods compute the deformation field usually less than a second (after training). The CNN-based methods can be categorized as supervised and unsupervised. The supervised methods require the ground-truth information in the dataset, while the ground-truth deformation fields are hard to obtain \cite{2017-supervisedRegistration,2018-supervisedRegistration}. Comparing with the supervised registration methods, unsupervised methods are not limited to the ground-truth information. According to the output of a methods, the registration methods can be divided into two categories: the displacement field registration and the diffeomorphic registration. The diffeomorphic methods compute the diffeomorphic deformation field to guarantee the desirable diffeomorphic property \cite{2020-SYM,2019-PVM,2020-diff-DeepFlash,IJCAI-reg1}. The displacement field methods output the deformation field directly from their CNN model, which directly use the deformation field to warp the moving image toward the fixed image \cite{2018-VM,2021-vitV}. Some recent studies employ the Transformer at the bottom of their network to overcome CNN's local receptive field shortcoming \cite{2021-vitV,2021-zhang-DTN}. The reason for placing the Transformer at the bottom of their networks is that the memory and computational complexity significantly increase with the higher resolution level. Motivated by the latest researches \cite{2021-CVT,2021-ResT}, we propose the CEMSA. Based on CEMSA, we build the CEMSA-Transformer-based symmetric network consisting of a total of ten Transformer blocks in 1/4, 1/8, and 1/16 resolution levels to enhance the contribution of transformers.


\section{Methods}
\begin{figure}
    \centering
    \includegraphics[width=0.7\textwidth,scale=0.8]{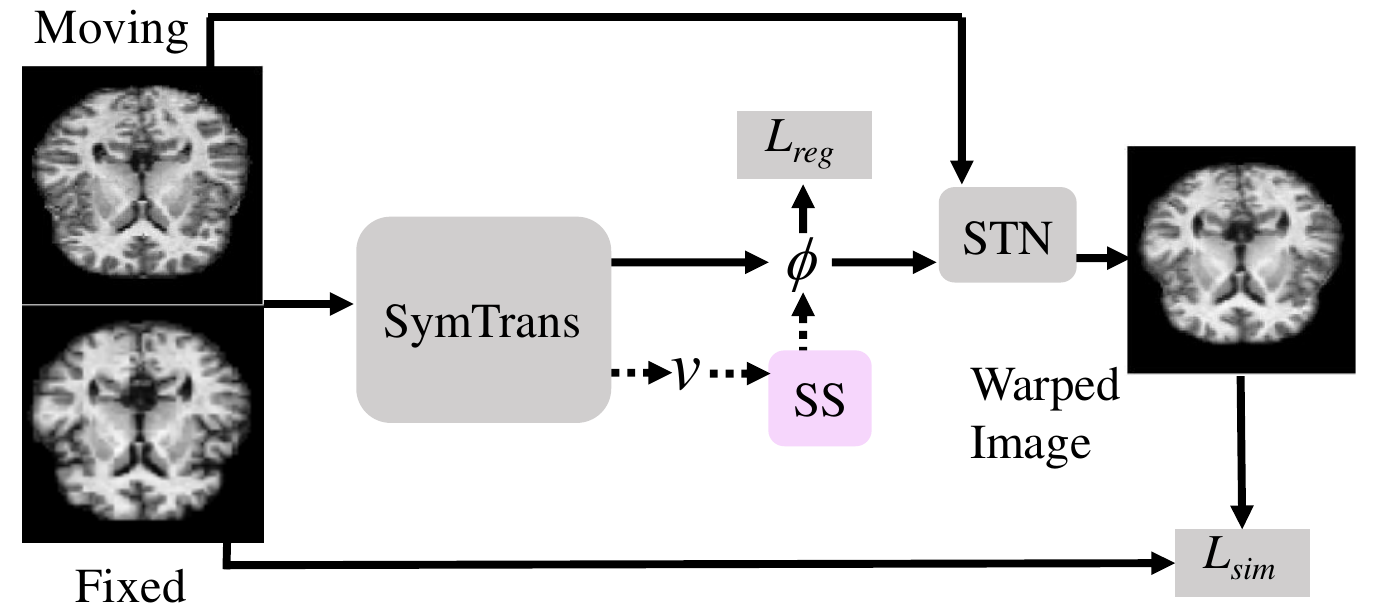}
    \caption{Overview of the proposed method for deformable image registration. The pink block named SS represents the scaling-and-squaring module. STN represents the spatial transform network. The dotted line indicates the workflow for diffeomorphic registration.}
    \label{fig:workflow}
\end{figure}
Let a pair of images be defined in the spatial domain ${\Omega}\subset{\mathcal{R}}^n, (n=3)$. Fig. \ref{fig:workflow} illustrates the overall architectures of deformable image registration in this paper. Briefly, the moving and fixed images (respectively denoted as $I_f$ and $I_m$) first input the proposed Transformer-based network, and then the network outputs the deformation field. Finally, the spatial transformation network \cite{2015-STN} is utilized to warp the moving image toward a fixed image via the deformation field. $\mathcal{L}_{sim}$ is the similarity loss function to evaluate the similarity between warped and fixed images. $\mathcal{L}_{reg}$ is the regularization to enforce the magnitude of the deformation field. 

\subsection{Efficient Transformer Block}
\begin{figure}
    \centering
    \includegraphics[width=0.7\textwidth,scale=0.8]{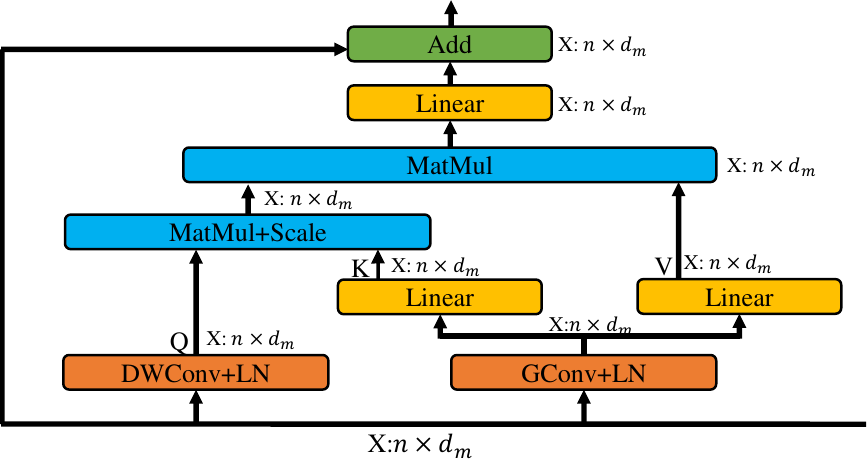}
    \caption{The proposed CEMSA block.}
    \label{fig:fig2}
\end{figure}
The standard Transformer usually takes up a lot of memory because Transformer has a large number of parameters, especially when applied in the 3D image tasks. To build the Transformer blocks symmetrically both in the encoder and decoder, we present a novel convolution-based efficient multi-head self-attention (CEMSA) for the Transformer block in this paper. The proposed CEMSA is shown in Fig. \ref{fig:fig2}. Compared with the standard Transformer, we employ the depth-wise separable and grouped convolution in the proposed CEMSA, which can further capture local spatial context, and reduce the semantic ambiguity and the computation costs. Each token input for attention function of \textbf{Q}, \textbf{K}, \textbf{V} can be summarily formulated as:
\begin{equation}
    \begin{aligned}
        x^{q, k, v}=\text{Flatten}(\text{Conv3D}(\text{Reshape}(x),s)),
    \end{aligned}
    \label{eq:Q}
\end{equation}
where $x$ is the input tokens to the CEMSA. DWConv is the depth-wise convolutional operation with the kernel size of $s$. GConv is the grouped convolutional operation with the number of the groups of the input's dimensions. After the DWConv and GConv, the LN (layer normalization) is applied. Then, two linear projection sets are utilized to obtain $\mathbf{K}$ and $\mathbf{V}$. After that, we adopt Eq. \ref{eq:MSA} to compute the attention function on \textbf{Q}, \textbf{K}, \textbf{V}. We use different $s$ for the depth-wise convolutional operation at 1/4, 1/8, 1/16 resolution levels. Then, we take advantage of the standard FFN to project the output of CEMSA. Thus, an CEMSA-based Transformer block is constructed.

It is worth noting that compared with the Standard and efficient MSA block mentioned and proposed in \cite{2020-CNNLimitation-VIT,2021-ResT}, we remove the position embedding to reduce the parameters further. \cite{2021-CVT} illustrates that the Transformer with the convolutional projection does not require position embedding because the convolutional projection represents the continuous positional information between tokens. To weaken the affection of eliminating the position embedding, and fully guaranteeing the positional information of each token, we use the single DWConv to compute the attention function on Q to get the spatial positional information. Compared to the existing efficient MSA approach \cite{2021-ResT}, we do not use the linear projection after the DWConv to maintain the positional information of each token. The role of the GConv operation is to reduce the parameters before the linear projections. According to the set number of groups $g$, GConv can reduce the number of parameters to $1/g$. We adopt Eq. \ref{eq:MSA} to compute the attention on Q, K, and V. As a result, the proposed CEMSA takes into account the spatial positional relations and the information of tokens at different positions.

\subsection{Symmetric Transformer-based Network}
\begin{figure*}
    \centering
    \includegraphics[scale=.6]{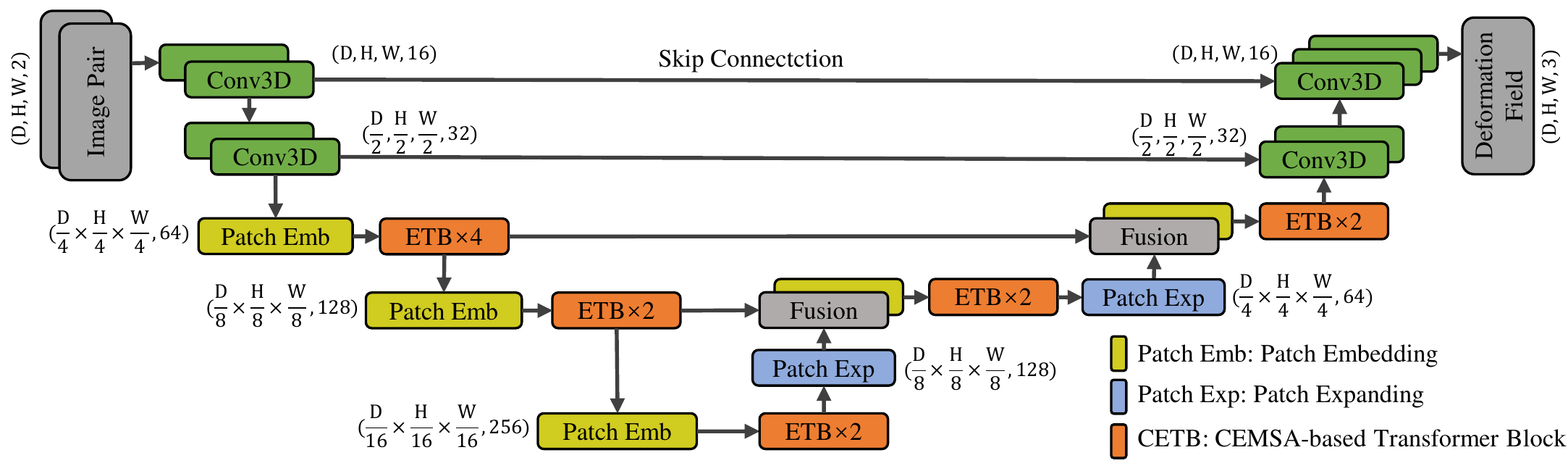}
    \caption{The propsed symmetric Transformer-based netowrk.}
    \label{fig:arch}
\end{figure*}

Using the proposed CEMSA-based Transformer, we can build the CEMSA-based Transformer blocks (SymTrans) both in encoder and decoder. The proposed SymTrans is shown in Fig. \ref{fig:arch}. The SymTrans is a U-shaped model like U-Net, consisting of 2 CNN-based encoding-decoding layers and 3 Transformer-based encoding-decoding layers. Each of the Transformer-based encoding-decoding blocks requires a sequence input. We utilize the convolutional operations to perform the \emph{Patch Embedding} operations before each Transformer in the encoder, with the stride of 2, kernel size of 3 (i.e., the patch size), to obtain the patch sequences with overlap. Before the feature maps input the next level Transformer block in the decoder, we utilize the \emph{Patch Expanding} operations to enlarge the feature maps. 

In detail, the \emph{Patch Expanding} operations consists of two linear projections, first expanding the size of feature maps by the factor of $2^3$, then expanding the feature maps dimension by the factor of 2. In the gap of SymTrans, the skip connections are used to concatenate the output feature maps from the Transformer in the encoder and the expanded feature maps from the decoder. Then, the \emph{Fusion} operations are utilized to reshape these two sequence feature maps to image form, then fuse them using the convolutional operations. At the half and original size resolution level, we utilize the convolutional blocks with the kernel size of 3, the stride of 1 (to the same resolution level), the stride of 2 (to the next resolution level) for encoding and decoding.

\subsection{Registration and Learning}
\subsubsection{Registration}
In this paper, we apply the SymTrans to displacement field registration and diffeomorphic registration. As shown in Fig. \ref{fig:workflow}, the deformation field can be generated in two ways to register a pair of images: the solid line following the SymTrans indicates the displacement field registration, the dotted line following the SymTrans indicates the diffeomorphic registration. The diffeomorphic branch ensures the diffeomorphism in registration. The diffeomorphism is a continuous, invertible, and one-to-one mapping. To achieve that, we follow \cite{2019-PVM,2020-SYM} and use the stationary velocity field with the efficient scaling-and-squaring approach to obtain the diffeomorphic deformation field. In the scaling-and-squaring approach, the deformation field is represented as a Lie algebra member that is exponentiated to generate the deformation field at time 1, which is a member of the Lie group, can be written as $\phi^{(1)}={\exp}(v)$. Starting from the initial deformation field at time 0, i.e., the output velo  city field from the SymTrans, can be formulated as:
\begin{equation}
    \begin{aligned}
        \phi^{(1/2^T)}=p+\frac{v(p)}{2^T},
    \end{aligned}
\end{equation}
where \emph{p} is the map of spatial locations. The recurrence to obtain the deformation field at time 1 can be written as:
\begin{equation}
    \begin{aligned}
        \phi^{(1/2^{t-1})}=\phi^{(1/2t)}\circ\phi^{(1/2t)}.
    \end{aligned}
\end{equation}
Hence, the time 1 deformation field $\phi^{(1)}=\phi^{(1/2)}\circ\phi^{(1/2)}$ is obtained.

\subsubsection{Learning}
The proposed SymTrans is optimized in an unsupervised manner by evaluating the similarity between aligned and fixed image. As shown in Fig. \ref{fig:workflow}, given a image pair ($I_m$,$I_f$), the Symtrans estimates the deformation field $\phi$. Then, the STN warps $I_f$ to obtain the warped image $\hat{I}_m$ (denoted as $\hat{I}_m={I_m}\circ{\phi})$. We apply the $L_2$ loss both on the registration similarity and smooth regularization. The loss function is defined as Eq. \ref{eq:energy} and formulated as $L={L_{sim}}(I_f,\hat{I}_m)+{\lambda}{L_{reg}}(\nabla\phi)$. We optimize the parameters of SymTrans by minimizing this loss function.

\section{Experiments} 
\subsection{Dataset and Metrics}
We demonstrate the proposed method on the task of brain MRI registration. We use the publicly available dataset OASIS, consisting of 425 T1-weighted brain MRI scans \cite{OASIS}, and 270 scans are selected in this dataset for our experiment. We first resample each scan to $256\times256\times256$ with the isotropic voxels size of $1mm\times1mm\times1mm$. Then, we conduct the standard preprocessing operation to normalize, affine transformation, and strip the skull using FreeSurfer \cite{FreeSurfer}. The segmentation maps of each scan viewed as ground truth for evaluation also is obtained through FreeSurfer. Each scan is cropped to $160\times192\times224$, then resampled to $96\times112\times96$. The dataset is split into 200, 34, and 36 scans for train, validation, and test sets, respectively. We sequentially, without repetition, combine two scans in the training set to obtain 39,800 permutations of image pairs. These scan pairs are used for training our proposed, and the baseline approaches. We conduct the basis atlas-based registration on the test set. Six and thirty scans are selected randomly as the atlas and moving images, respectively. 

Baseline methods and proposed methods are evaluated using the Dice similarity coefficients (DSC), which calculates the overlapping between the ground truth segmentation maps and the warped moving image corresponding segmentation maps. We count the negative Jacobian determinant ${|J(\phi)|}\leq0$ to denote the number of folding. ${|J(\phi)|}\leq0$ relates where the voxels lose topology preservation and the violate the diffeomorphic property when transformed via the deformation field.

\subsection{Baseline Methods}
We compare the proposed method SymTrans with five approaches, including one traditional and four deep-learning methods. The symmetric image normalization registration method (SyN) is a traditional iterative method to compute the deformation field \cite{SyN}. We use the SyN implementation in the ANTs \cite{Ants} toolbox and set the iteration to [100,100,100]. The deep learning baseline methods, including the CNN-based VoxelMorph \cite{2018-VM}, the CNN-based SYMNet \cite{2020-SYM}, the Transformer-based ViT-V-Net \cite{2021-vitV} and the Swin-Transformer-based TransMorph \cite{transmorph}. We use the publicly available implementation of these four deep learning methods. We train the VoxelMorph, SYMNet, Vit-V-Net and TransMorph with the setting of their suggested hyperparameters, on the same data set splitting, respectively.

\subsection{Implementation Details}
The proposed framework is implemented by using the PyTorch. The STN in our method is the same as the one utilized in VoxelMorph , Vit-V-Net, and TransMorph. We set the regularizing parameter $\lambda$ to 0.02. We employ the Adam optimizer to optimize the parameters of the proposed network, with a learning rate of 1e-4, on an NVIDIA RTX3080 10 GB GPU. The maximum iterations of training for the deep-learning approaches are 300k. 

The detailed configures of the proposed CEMSA-based Transformer during training is following: $s=\{24,16,12\}$ at 1/4, 1/8, 1/16 resolution stages; the number of heads is $\{2,4,8\}$ at each resolution stage; the patch size is $\{3,3,3\}$; the number of the grouped convolution's groups is equal to the input embedding dimension.

\subsection{Results}

\subsubsection{Registration Accuracy}

\begin{table}
  \centering
  \setlength{\abovecaptionskip}{0.5cm}
  \setlength{\tabcolsep}{7mm}{
    \begin{tabular}{cccc}
    \toprule
    Method & DSC & ${|J(\phi)|}\leq0$ \\  
    \midrule
    Affine &  0.520 (0.058)    &  -\\
    SyN & 0.662 (0.038)    & 40.683 (78.042) \\
    VoxelMorph    & 0.726 (0.031)     & 1453.778 (624.714) \\
    SYMNet &0.719 (0.025)       & 1205.789 (365.011)   \\
    Vit-V-Net & 0.730 (0.031)   & 1563.088 (631.037)  \\
    TransMorph & 0.742 (0.027) & 1631.978 (574.568) \\
    SymTrans & \textbf{0.747 (0.026)}    & 1581.033 (587.560) \\
    diff-SymTrans & \textbf{0.742 (0.025)}     & \textbf{2.033 (9.942)} \\

    \bottomrule
    \end{tabular}%
    }
      \caption{Qualitative comparison between our frameworks and baseline methods. DSC higher is better, and ${|J(\phi)|}\leq0$ lower is better. Standard deviations are in bracket.}
  \label{tab:acc}%
\end{table}%

\begin{figure}
    \centering
    \includegraphics[width=0.8\textwidth,scale=1.0]{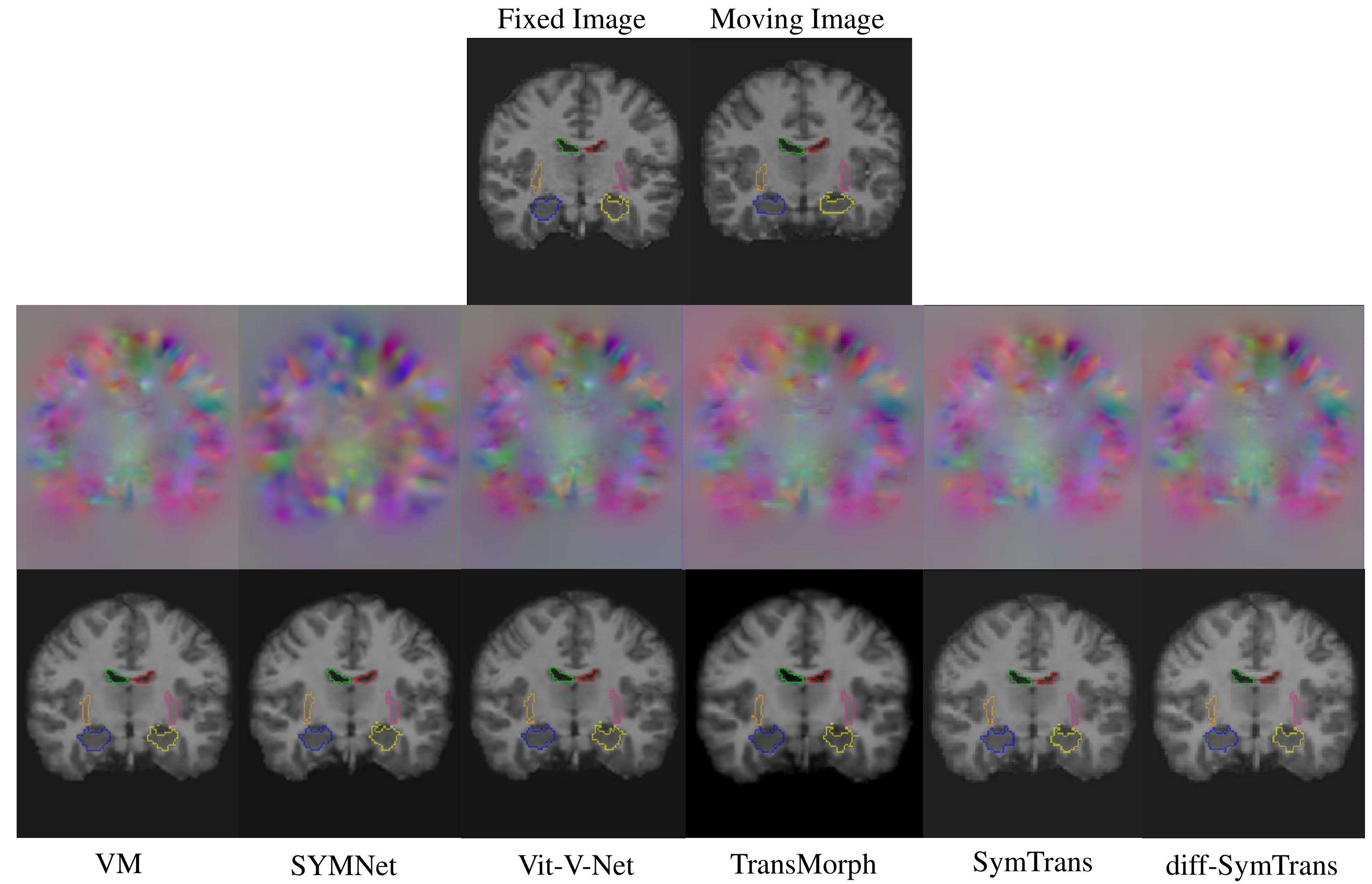}
    \caption{The atlas-based registration of lateral-ventricle, thalamus, and hippocampus by the VoxelMorph, SYMNet, Vit-V-Net, TransMorph, and the proposed SymTrans and diff-SymTrans.}
    \label{fig:boundries}
\end{figure}

Fig. \ref{fig:boundries} shows the registration results of a pair of images. The boundaries of three segmentation maps are marked in the sampled slices to observe the deformation of each anatomical structure. We quantitatively evaluate the accuracy of the baseline methods and the proposed SymTrans using the DSC metric. The non-positive Jacobian determinants are utilized to assess the number of folding. Table \ref{tab:acc} shows the results of different methods on the same test set. The proposed SymTrans, applied to displacement field registration, produces the highest average DSC than the baseline methods. The diffeomorphic registration using SymTrans (denoted as diff-SymTrans) still gives the higher average DSC than baseline methods and decreases the average number of folding much lower, which guarantees the topology of the original moving image. Besides, the lower standard deviations of the SymTrans and diff-SymTrans show strong stability of the proposed SymTrans. 

\begin{figure*}
    \centering
    \includegraphics[width=1.0\textwidth]{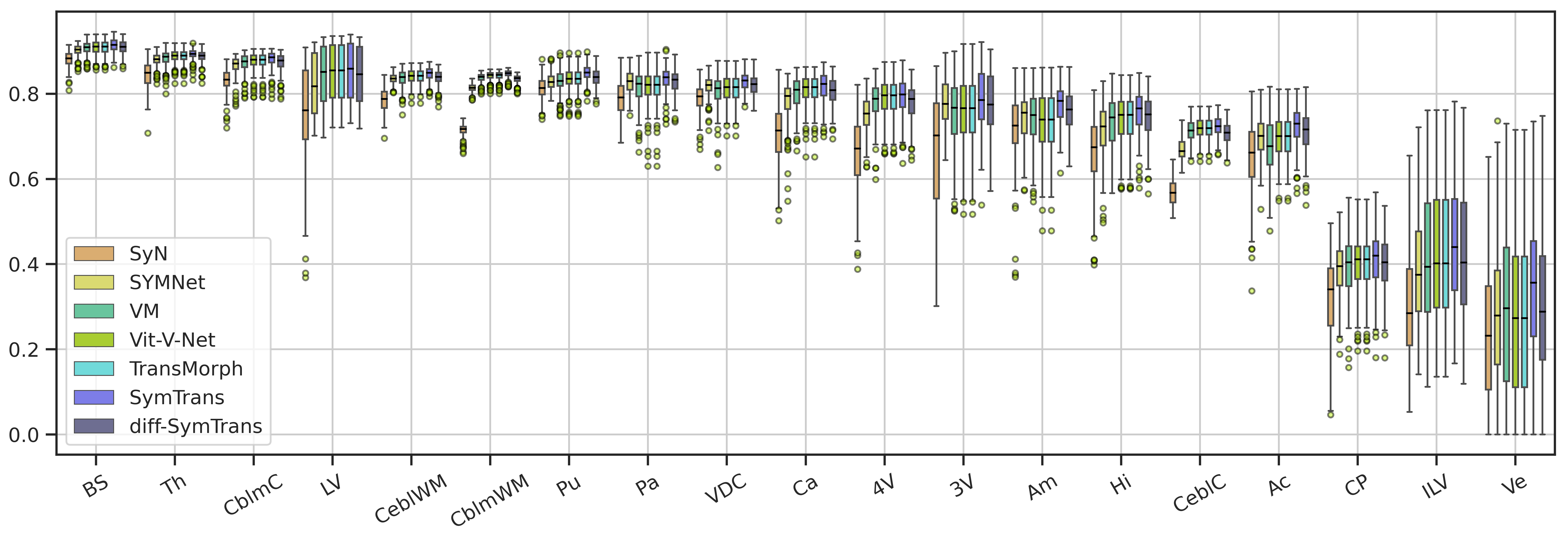}
    \caption{A boxplot illustrating the DSC of each anatomical structure segmentation for SyN, VoxelMorph, SYMNet, Vit-V-Net, and ours. We averaged the Dice values of the left and right brain hemispheres and combined them into one structure for visualization.}
    \label{fig:box}
\end{figure*}

To demonstrate the alignment results of each anatomical structure, we report the DSCs of 35 anatomical structures in Fig. \ref{fig:box}.The abbreviations in Fig. \ref{fig:box} are: Brainstem (BS), thalamus (Th), cerebellum cortex (CblmC), lateral ventricle (LV), putamen (Pu), pallidum (Pa), cerebral white matter (CeblWM), ventral DC (VDC), caudate (Ca), Amygdala (Am) hippocampus (Hi), 3rd ventricle (3V), 4th ventricle (4V), amygdala(Am), CSF (CSF), cerebral cortex (CeblC), inf-lateral ventricle (ILV), Vessel (Ve) and choroid plexus (CP). As we can see, the proposed SymTrans outperforms the compared registration approaches on all of the 19 combined structures. The diff-SymTrans yields better results than all baseline methods except TransMorph and SymTrans, while producing minimal folding. To sum up, the proposed symmetric Transformer model based on the CEMSA achieves the best results.

\subsubsection{Computational Complexity}

\begin{table}
  \centering
    \setlength{\abovecaptionskip}{0.5cm}
    \setlength{\tabcolsep}{7mm}{
    \begin{tabular}{cccc}
    \toprule
    Method & Trans. L. & Params (M) & FLOPs (G) \\
    \midrule
    VoxelMorph    & -    & 0.29  & 59.82  \\
    SYMNet & -   & 1.12  & 44.51  \\
    Vit-V-Net & 1/16   & 31.50   & 65.77 \\
    TransMorph & 1/4 & 46.69 & 112.75\\
    SymTrans & 1/4  & 16.05  & 63.53  \\
    \bottomrule
    \end{tabular}%
    }
      \caption{The parameters and FLOPs comparison of different methods for registration. Input image size is $96\times112\times96$ by default. Trans. L.: The starting deployment location of the Transformer.}
  \label{tab:param}%
\end{table}%

To illustrate the effectiveness of the proposed CEMSA, we compare its parameters and the FLOPs with the baseline approaches. Table \ref{tab:ab} shows the FLOPs and the parameters of each method. The CNN-based networks, both VoxelMorph and SYMNet, have fewer parameters and FLOPs than Transformer-based models because Transformer-based models have many linear operations, which enlarge the parameters and FLOPs scale. Among these three Transformer-based methods, our SymTrans achieves the lowest FLOPs. Compared with Vit-V-Net and TransMorph, the parameters of the SymTrans are much fewer, and the FLOPs are fewer than theirs. Specifically, Vit-V-Net employs 12 Transformer blocks at the bottom of their model, each block containing 1.76M parameters. TransMorph employs the Swin-Transformer blocks at the 1/4 resolution stage, which model the input embedding patches with the dimensions of 96. In the SymTrans encoder, the depth of the CEMSA-based Transformer at each resolution stage is equal to the depth of the Swin-Tranformer in TransMorph. 

In general, the parameters are gained while the size of the input token raises. Symtrans applies the CEMSA-based Transformer blocks at 1/4, 1/8, and 1/16 resolution levels in the framework. Even applying the CEMSA-based Transformer blocks at so manyresolution stages, SymTrans has about 49\% fewer parameters than Vit-V-Net, and 67\% fewer parameters than TransMorph. In practice, the GPU memory occupied during training is about 3 GB with a batch size of 1 and an input image size of $96\times112\times96$ on our server. Vit-V-Net and Transmorph occupy about 6 GB and 7 GB of GPU memory with the input padded image size of $96\times128\times96$. Statistical results of parameters and FLOPs indicate that the proposed CEMSA is feasible to reduce parameters, which provides a basis to apply the Transformer at the high-resolution levels.

\subsubsection{Ablation Studies}
\begin{table}
  \centering
    \setlength{\abovecaptionskip}{0.5cm}
    \setlength{\tabcolsep}{4mm}{
    \begin{tabular}{ccccc}
    \toprule
    Method & \multicolumn{1}{l}{E-SymTrans} & \multicolumn{1}{l}{D-SymTrans} & \multicolumn{1}{l}{B-SymTrans} & \multicolumn{1}{l}{SymTrans} \\
    \midrule
    DSC  &0.734 (0.028)  & 0.717 (0.033)  & 0.714 (0.034) &  0.740 (0.027) \\
    \bottomrule
    \end{tabular}%
    }
      \caption{Comparison of placing the CEMSA-based Transformer in different branch of the proposed network. Standard deviations are in bracket.}
  \label{tab:ab}%
\end{table}%

We investigate the performance when the CEMSA-based Transformer is applied at different locations in the network to demonstrate that the symmetric framework is effective. The original Symtrans and all the ablation are utilized to perform the displacement field registration. We train the ablation variants for 100k iterations. Then, we find the best weights on the validation set and test these variants on the test set. 

Table \ref{tab:ab} reports the DSC results of three variant SymTrans. E-SymTrans contains the CEMSA-based Transformer blocks in the encoder and replaced the CEMSA-based Transformer blocks with the convolutional blocks in the decoder. D-SymTrans indicates that only the CEMSA-based Transformer blocks are utilized in the decoder, and the rest, as shown in Fig. \ref{fig:arch}, are convolutional blocks. \emph{Patch Embedding} and the \emph{Fusion} blocks in these two ablations are replaced with the basis convolutional blocks. B-SymTrans is the CNN-based architecture that applies 10 CEMSA-based Transformer blocks at the bottom. Each convolution block is followed by a LeakyReLU activation to construct a Conv block. The depths of Conv blocks are the same as the depths of the replaced CEMSA-based Transformer. \emph{Patch Expanding} blocks are replaced with the deconvolutional operation. The structures form of E-SymTrans and D-SymTrans correspond to the structures form of TransMorph and Vit-V-Net. We observe that the original SymTrans achieves the best performance. The results of these ablation variants identify that employing the CEMSA-based Transformer at the high-resolution levels of the network and applying them symmetrically as encoder and decoder enhance the registration accuracy. That demonstrates that modeling high-resolution feature maps with the symmetric architecture can facilitate the model to recognize meaningful semantic correspondences to anatomical structures.

\section{Conclusion}
This paper proposes an CEMSA mechanism to capture local spatial context, reduce semantic ambiguity and parameters. Based on the proposed CEMSA, we build the Symtrans for deformable image registration, which takes advantage of the long-range spatial relevance for feature enhancements. The Transformer blocks based on CEMSA are not only applied at the bottom but also at the higher-resolution levels both in encoder and decoder. The qualitative and quantitative evaluations demonstrate that the SymTrans promotes the semantically meaningful correspondence of anatomical structures and provide the state-of-the-art registration performance. Furthermore, the ablation studies illustrate the impact on performance when the Transformer is applied on different components (i.e., encoder and decoder) of the model, which indicates the effectiveness of symmetric scheme and the importance of building transformers at the high-resolution levels.

{\small
	\bibliographystyle{ieee_fullname}
	\bibliography{SymTrans}
}

\end{document}